\RequirePackage{fix-cm}
\documentclass[twocolumn]{svjour3}          
\smartqed  
\usepackage{graphicx}
\usepackage{natbib}
\usepackage[pagebackref=false,breaklinks=true,citecolor=blue,colorlinks,linkcolor=red,bookmarks=false,urlcolor=blue]{hyperref}

\usepackage{array}
\usepackage{multirow}
\newcolumntype{H}{>{\setbox0=\hbox\bgroup}c<{\egroup}@{}}
\usepackage{amsmath}
\usepackage{bm}
\usepackage{amsfonts}

\usepackage{hyperref}
\usepackage{graphicx}
\usepackage{amssymb}
\usepackage{algorithm}
\usepackage{algorithmic}

\usepackage[misc]{ifsym} 

\usepackage{bbm}
\usepackage{dsfont}
\usepackage{pifont}

\usepackage{microtype}

\usepackage{booktabs}

\usepackage{subfig}

\usepackage{xspace}
\newcommand{\onedot}{\ifx\@let@token.\else.\null\fi\xspace}

\begin{document}
\sloppy 

\title{Breaking the SSL-AL Barrier: A Synergistic Semi-Supervised Active Learning Framework for 3D Object Detection}

\author{Zengran Wang$^{1,2}$ \and Yanan Zhang$^{3}$ \and  Jiaxin Chen$^{2}$ \and Di Huang\textsuperscript{\rm 1,2,\Letter} 
}

\authorrunning{Zengran Wang {\it et al.}}
\titlerunning{A Synergistic Semi-Supervised Active Learning Framework for 3D Object Detection}

\institute{
Zengran Wang\\ \email{wangzr333@buaa.edu.cn}
\\~\\ 
Yanan Zhang\\ \email{yananzhang@hfut.edu.cn}
\\~\\ 
Jiaxin Chen\\ \email{jiaxinchen@buaa.edu.cn} 
\\~\\ 
Di Huang\\ \email{dhuang@buaa.edu.cn} \\~\\
$^1$ State Key Laboratory of Complex and Critical Software Environment, Beihang University, Beijing 100191, China \\~\\
$^2$ School of Computer Science and Engineering, Beihang University, Beijing 100191, China \\~\\
$^3$ School of Computer Science and Information Engineering, Hefei University of Technology, Hefei 230601, China
\\~\\
\textsuperscript{\rm \Letter} refers to the corresponding author.
}

\date{Received: date / Accepted: date}

\maketitle
\begin{abstract}
To address the annotation burden in LiDAR-based 3D object detection, active learning (AL) methods offer a promising solution. However, traditional active learning approaches solely rely on a small amount of labeled data to train an initial model for data selection, overlooking the potential of leveraging the abundance of unlabeled data. Recently, attempts to integrate semi-supervised learning (SSL) into AL with the goal of leveraging unlabeled data have faced challenges in effectively resolving the conflict between the two paradigms, resulting in less satisfactory performance. To tackle this conflict, we propose a \textbf{S}ynergistic \textbf{S}emi-\textbf{S}upervised \textbf{A}ctive \textbf{L}earning framework, dubbed as S-SSAL. Specifically, from the perspective of SSL, we propose a Collaborative PseudoScene Pre-training (CPSP) method that effectively learns from unlabeled data without introducing adverse effects. From the perspective of AL, we design a Collaborative Active Learning (CAL) method, which complements the uncertainty and diversity methods by model cascading. This allows us to fully exploit the potential of the CPSP pre-trained model. Extensive experiments conducted on KITTI and Waymo demonstrate the effectiveness of our S-SSAL framework. Notably, on the KITTI dataset, utilizing only 2\% labeled data, S-SSAL can achieve performance comparable to models trained on the full dataset. The code has been released at \href{https://github.com/LandDreamer/S\_SSAL}{https://github.com/LandDreamer/S\_SSAL}.

\end{abstract}    
\section{Introduction}

\begin{figure*}[ht]
\centering
\includegraphics[width=0.97\textwidth]{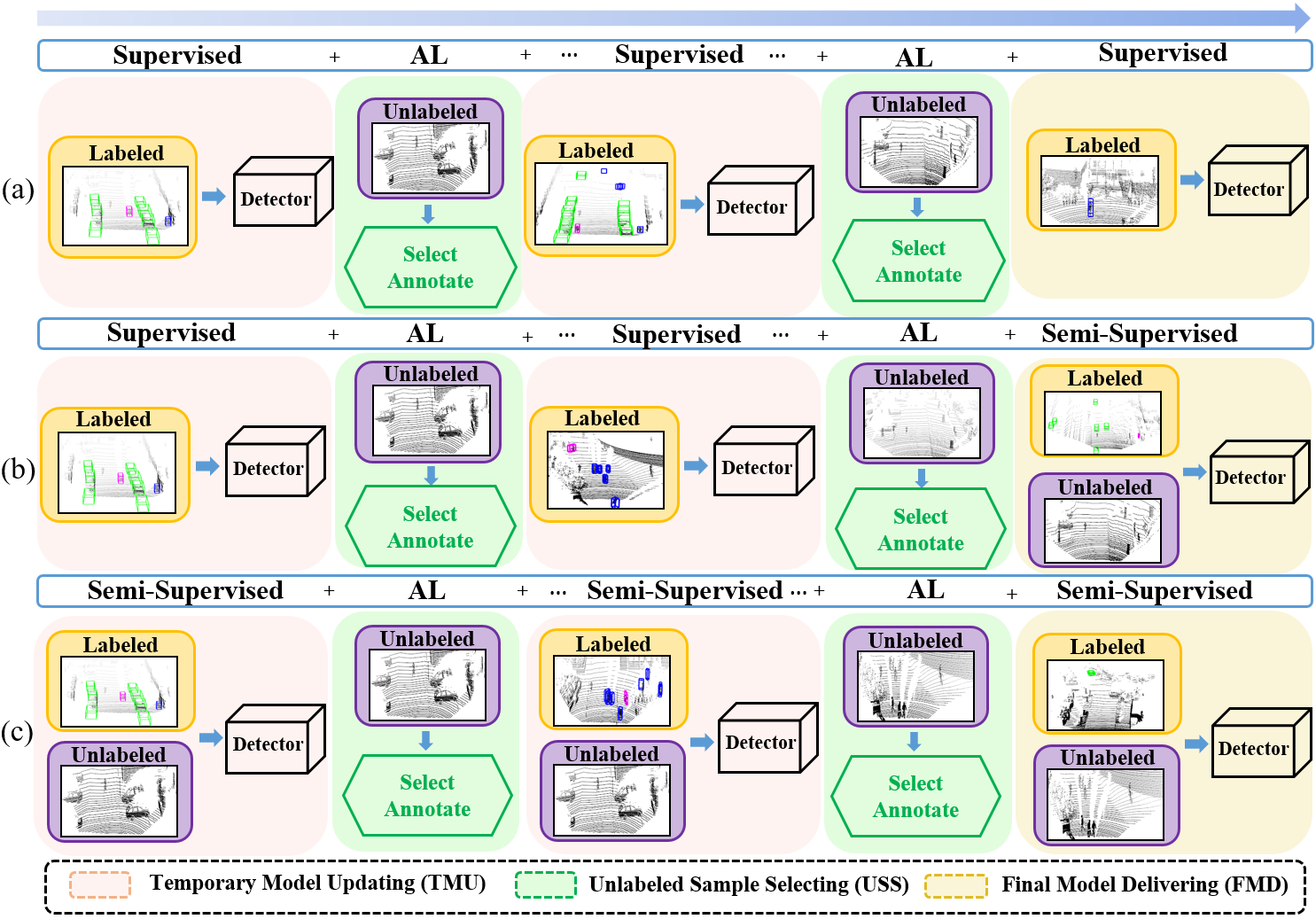} 
\caption{
The illustration depicts different paradigms for combining Active Learning (AL) and Supervised/Semi-supervised Learning (SL/SSL): (a) Solely utilizing SL in all stages. (b) Employing SSL only in the final model delivering stage. (c) Integrating SSL across all stages. Paradigm (c) achieves enhanced performance by incorporating unlabeled data compared to paradigm (a). However, traditional SSL methods face conflicts with AL in the temporary model updating stage, leading to suboptimal data selection. Thus, paradigm (c) performs less effectively than paradigm (b).
}
\label{fig:sslalbad}
\end{figure*}

Being a fundamental task in autonomous driving, LiDAR-based 3D object detection plays a crucial role in perceiving semantic and spatial clues, which recognizes and locates objects in 3D scenes based on input point clouds captured by LiDAR sensors. During the past few years, a large number of efforts~\citep{yan2018second, zhou2018voxelnet, lang2019pointpillars, chen2022sasa, yang20203dssd, shi2020pv, shi2023pv} have been made with the performance of major public benchmarks~\citep{kitti, nuscenes, waymo} rapidly and consistently increasing. Unfortunately, current methods are deep learning based, substantially dependent on labeled data. For instance, the Waymo dataset ~\citep{waymo} alone encompasses over 10 million ground-truth (GT) 3D boxes. The labor-intensive and time-consuming nature of annotating extensive datasets creates a bottleneck, hindering the advancement in this field.

Active learning (AL)~\citep{scaleal, deepal, deep3dal} offers a promising solution to overcoming this drawback. It selects a small subset from all samples as the most informative data to measure the benefits of a fully annotated dataset. By adaptively choosing ``good" samples to label, AL significantly reduces the burden of data acquisition and annotation and shows the potential to facilitate LiDAR-based 3D object detection~\citep{crb, kecor, rareal, ad3dal}.

The AL paradigm typically consists of three phases, \emph{i.e.} (1) temporary model updating (\textbf{TMU}), (2) unlabeled sample selecting (\textbf{USS}), and (3) final model delivering (\textbf{FMD}). In TMU, a temporary model is built or enhanced with the set of available labeled data, which is further applied to generate pseudo annotations; in USS, some unlabeled data are screened out according to certain criteria and annotated by the temporary model obtained; and in FMD, the final model is output. In general, TMU and USS are jointly conducted for multiple iterations while FMD operates once in the end. Traditional AL methods only make use of labeled data, as shown in Fig.~\ref{fig:sslalbad} (a). Since the large amount of unlabeled data conveys rich information, which helps better understand the distribution of all data rather than that of labeled ones, overlooking them leaves much room for improvement.

With the progress achieved in semi-supervised learning (SSL), some preliminary attempts~\citep{boxal, alwod, activeteachers} have been made to integrate such techniques in AL, where SSL contributes to the performance gain by assigning pseudo-labels to unlabeled data based on the prediction of the model trained on label data~\citep{sess, 3dioumatch}. As depicted in Fig.~\ref{fig:sslalbad} (c), they employ SSL to strengthen both the temporary and final model in the TMU and USS phases and demonstrate that the semi-supervised active learning (SSAL) paradigm is superior to the traditional one (Fig.~\ref{fig:sslalbad} (a)). However, the combination of SSL and AL is not as straightforward as they expect, in particular for the synergy of TMU and USS. As we know, uncertainty-based metrics are widely adopted in AL and the samples with higher uncertainties are more likely to be selected for annotation in USS. On the other side, the samples of higher uncertainties may suffer from low confidence scores due to the instability of SSL in TMU. In this case, a conflict arises, where the assignment of incorrect pseudo labels to objects in SSL inevitably becomes a significant source of noise and makes AL struggle to accurately assess their uncertainties. As this iterates for multiple rounds, current SSAL is prone to converge with sub-optimal results, even inferior to that of a degraded paradigm only applying SSL in FMD (Fig.~\ref{fig:sslalbad} (b)).

To tackle the conflict between SSL and AL, we propose a synergistic semi-supervised active learning (S-SSAL) framework. 
\textbf{From the perspective of SSL}, in TMU, we present a method, namely collaborative pseudo-scene pre-training (CPSP), to effectively leverage unlabeled data while bypassing the side effects aforementioned. The main idea is to selectively learn only from confident objects. To this end, we generate pseudo-scenes of unlabeled data using confident objects with the ones of high uncertainties excluded. Pre-training on these pseudo-scenes thus ensures that unconfident objects are not disturbed by their own pseudo-labels, which largely mitigates the negative impact of mislabeling and noise.
\textbf{From the perspective of AL}, we propose a novel AL approach that leverages our CPSP pre-trained model to its fullest potential. Our method introduces an ensemble-based strategy to enhance the reliability of uncertainty measures, addressing the challenge of pre-trained models ``forgetting" confident objects. Additionally, we employ a similarity-based technique to reduce redundancy and mitigate class imbalance, especially in outdoor LiDAR scenes, where long-tailed distributions and mixed-class objects complicate the active learning process.

In summary, our contributions are as follows:
\begin{itemize}

  \item We point out the conflict between AL and SSL and propose a novel SSAL framework to address it, where the CPSP method is presented to effectively leverage unlabeled data to facilitate model training.

  \item   We propose a novel AL approach that maximizes the potential of the CPSP pre-trained model with an ensemble-based strategy to enhance uncertainty measures, a diversity technique to reduce redundancy, and make ensure class balance sampling.

  \item We do extensive experiments on the KITTI and Waymo datasets and reach state-of-the-art results. Especially, on KITTI, we use only 2\% labeled data and achieve comparable performance to the model trained on the full set.

\end{itemize}
\section{Related Work}
\subsection{LiDAR-based 3D Object Detection.}
With the rapid advancement of LiDAR technology, significant progress has been made in 3D object detection. Existing approaches can be broadly categorized based on their representation strategies for LiDAR point clouds into three types: Point-based, Voxel-based, and Point-voxel-based methods.
Point-based methods~\citep{shi2019pointrcnn,yang20203dssd,zhang2021pc,zhang2022cat} directly operating on raw point clouds. These methods wholly preserve the geometry information of a point cloud but have relatively higher latency. Voxel-based methods~\citep{zhou2018voxelnet,yan2018second,zhou2023octr,zhang2024stal3d,zhang2024cmae,zhang2023sa,zhang2024geobev,jiang2025fsd} voxelize point clouds into 2D/3D compact grids and then collapse it to a bird's-eye-view representation. They are computationally effective but voxelization inevitably introduces quantization errors. Point-Voxel-based methods~\citep{yang2019std,shi2020pv,he2020structure,shi2023pv} integrate the advantages of both Point-based methods and  Voxel-based methods together. However, current 3D object detection methods heavily rely on fully annotated large-scale point cloud datasets, resulting in high labor and cost for annotation. In this paper, we are committed to using 3D active learning methods to select the most valuable data for annotation, rather than directly annotating it in full, while hoping to achieve comparable perceptual performance.

\begin{figure*}[ht]
\centering
\includegraphics[width=0.999\textwidth]{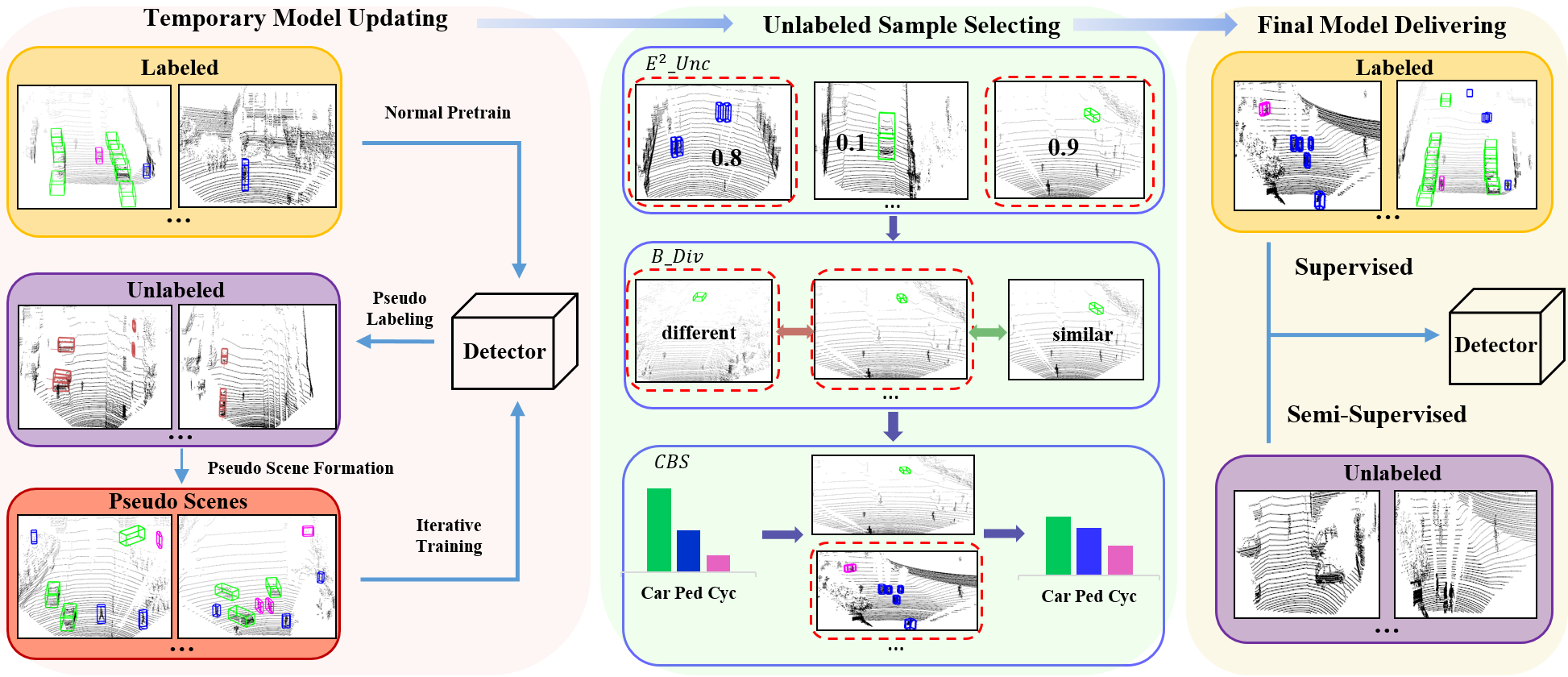} 
\caption{
Overview of our S-SSAL framework. In the Temporary Model Updating stage(TMU), we propose creating pseudo scenes with confident objects for model pre-training (CPSP). 
Subsequently, in the Unlabeled Sample Selecting stage(USS), we design a collaborative active learning method to select valuable data for annotation (CAL). 
Finally, in the  Final Model Delivering stage(FMD), we leverage traditional semi-supervised learning methods to enhance the model performance.
}
\label{fig1}
\end{figure*}

\subsection{Active Learning}

Active learning methods have gained significant attention in various domains to alleviate the labeling burden. These methods can be broadly categorized into two main types: uncertainty-based~\citep{bald, DBAL} and diversity-based approaches~\citep{clusteringal, coreset, cdal}. 
Uncertainty-based methods leverage uncertainty to identify informative samples for annotation while diversity-based methods prioritize capturing the diversity and representativeness of the dataset.
Furthermore, recent research~\citep{alqr, badge} has explored the integration of uncertainty-based and diversity-based approaches to leverage the advantages of both. 

Recently, there has been increased interest in applying active learning to object detection tasks. Unlike image classification, active learning for object detection presents unique challenges due to the complexities of localizing and identifying objects within images. One approach, MI-AOD~\citep{MIAOD} treats unlabeled images as bags of instances, using adversarial classifiers to measure uncertainty. AL-MDN~\citep{ALMDN} utilizes mixture density networks for probabilistic outputs, while ENMS~\citep{enms} applies entropy-based non-maximum suppression to assess uncertainty. PPAL~\citep{ppal} offers a plug-and-play active learning method. 
However, active learning for LiDAR-based object detection needs further research due to the differences between images and outdoor LiDAR scenes. Some recent studies have begun to tackle this issue. For example, CRB~\citep{crb} focuses on filtering redundant 3D bounding box labels based on conciseness, representativeness, and geometric balance. KECOR~\citep{kecor} presents a novel strategy called kernel coding rate maximization to identify the most informative point clouds for labeling. However, these methods may struggle with class imbalance caused by long-tailed distributions in outdoor scenes and do not effectively utilize available unlabeled data.

\subsection{Semi-Supervised Active learning}
Semi-supervised learning (SSL) techniques~\citep{sslod1, sslod2, sess, 3dioumatch, ProficientTeachers, hssda, dqs3d} aim to enhance model performance by leveraging abundant unlabeled data. These methods can be integrated with active learning (AL) to further optimize data annotation efforts~\citep{Notalllabels, activeteachers, boxal, alwod, joint3dalssl}.
In most frameworks, SSL is employed for model pre-training during the Temporary Model Updating (TMU) stage, after which AL identifies the most informative samples for annotation. However, many approaches overlook the potential conflicts between SSL and AL during TMU. These methods often rely on pseudo-labeling techniques that may degrade performance due to the noises~\citep{activeteachers, boxal, alwod}.
On the other hand, Joint3D~\citep{joint3dalssl} mainly uses consistency loss, which is less affected by conflicts between SSL and AL, but it still lacks sufficient support for effective semi-supervised learning in 3D object detection. Similarly, NAL~\citep{Notalllabels} uses an auto-labeling scheme to reduce distribution drift. However, this method relies on a specific loss function, making it difficult to supervise established 3D detectors. As a result, it may struggle to learn effectively from unlabeled data in 3D detection tasks.
In this paper, we propose a bidirectional collaborative semi-supervised active learning framework, which addresses the conflicts between SSL and AL, effectively unleashing the potential of unlabeled data for 3D object detection.

\section{Method}

\subsection{Framework Overview}
As illustrated in Fig.~\ref{fig1}, our \textbf{S}ynergistic  \textbf{S}emi-\textbf{S}upervised \textbf{A}ctive \textbf{L}earning framework (S-SSAL) consists of three main components: Temporary Model Updating (TMU), Unlabeled Sample Selecting (USS), and Final Model Delivering (FMD).
\textbf{In the TMU stage}, we initiate the process with normal pre-training, where a small set of randomly sampled data is used to train the initial model. Subsequently, our Collaborative PseudoScene Pre-training tailored for active learning is performed, creating pseudo scenes with confident boxes to enhance the model performance.
\textbf{In the USS stage}, we employ the innovative Collaborative Active Learning method, which entails the strategic selection of informative data from the unlabeled pool, empowering the model to concentrate on challenging instances.
\textbf{In the FMD stage}, semi-supervised learning is conducted to further refine the model performance, which utilizes both labeled and unlabeled data to train the model, capitalizing on insights gained from the active learning process. It's important to note that our framework is compatible with various existing semi-supervised methods, providing flexibility in choosing the most suitable approach.

\subsection{Collaborative PseudoScene Pre-training}
\begin{figure*}[ht]
\centering
\includegraphics[width=0.9\textwidth]{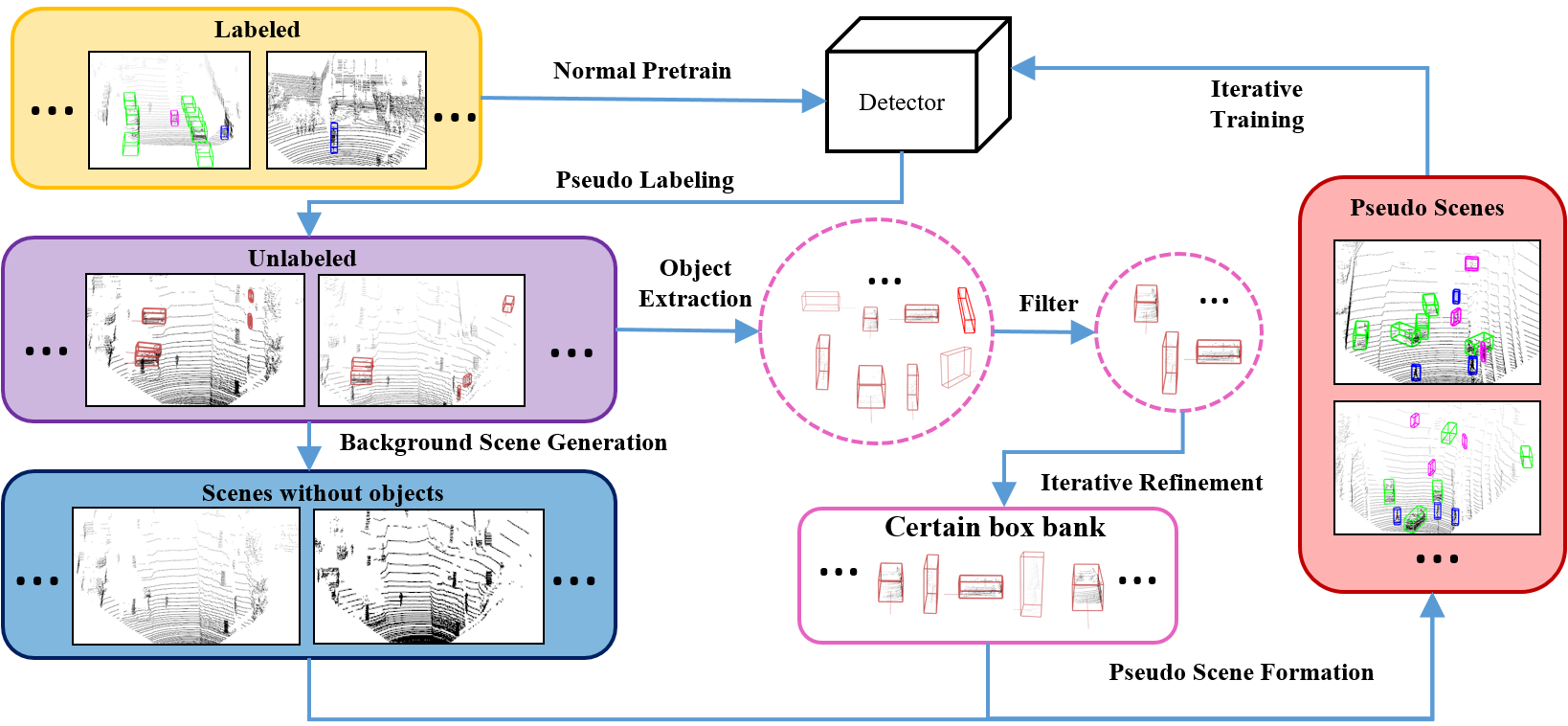} 
\caption{
The illustration of the Collaborative PseudoScene Pre-training (CPSP) module. We extract confident objects from unlabeled scenes based on their uncertainty and store them in a box bank, which is iteratively updated to maintain its quality. Additionally, we remove point clouds corresponding to the predicted boxes, creating "background" scenes without any objects. The point cloud from the box bank is then inserted into these "background" scenes, forming pseudo scenes.
}
\label{fig:CPSP}
\end{figure*}

To meet the requirements of active learning, we propose the Collaborative PseudoScene Pre-training(CPSP) approach, which is specifically designed to support active learning by creating pseudo scenes that focus on confident objects while excluding unconfident boxes.
The entire process is illustrated in Fig.~\ref{fig:CPSP}.

\subsubsection{Confident Object Extraction}
To optimize the extraction of confident objects for model pre-training, we employ a multi-step approach that enhances the quality of our training data. Initially, we utilize a Confident Object Filtering module to extract confident objects from unlabeled scenes, providing crucial information for model training. These extracted objects are then stored in a box bank to preserve and manage the collected object information. Furthermore, we incorporate an Iterative Refinement mechanism that iteratively generates confident boxes from the unlabeled data, integrating them into the box bank to create high-quality pseudo labels for the models. This process is essential for improving the robustness and accuracy of the model.

\textbf{Confident Object Filtering.}
To ensure compatibility with active learning(AL), we utilize the same uncertainty measure employed in the AL process. By applying this uncertainty measure to the objects, we collect their uncertainty scores and employ clustering techniques on these scores to identify the group with the lowest uncertainty scores. Traditional semi-supervised learning (SSL) methods typically use fixed thresholds or top-k selections to filter confident objects. However, we find that score distributions vary greatly across classes and models, making it difficult to set an appropriate class-specific threshold. Additionally, model performance fluctuations complicate the selection of a consistent top-k value—if $k$ is too small, too few objects are selected; if $k$ is too large, noise increases. To address this, we leverage clustering techniques, such as KMeans~\citep{hssda}, on uncertainty scores to select confident objects. Clustering groups objects with similar patterns, and by choosing the highest number of cluster centers, we enhance the reliability of confident object selection. Once confident groups are identified, we filter objects within these groups for selection. The object information is represented as $O=\{cls, loc, score, scene_{id}, pc\}$, where $cls$ denotes the class label, $loc \in \mathbb{R}^{7}$ represents the object’s location and orientation, $score \in \mathbb{R}^1$ is the uncertainty score, $scene_{id}$ is the scene index, and $pc \in \mathbb{R}^{n \times 3}$ contains the point cloud of the object. We also extract background objects likely to be false positives, as done in~\citep{fp3d}. All extracted information is stored in a box bank for easy access and management.

\textbf{Iterative Refinement.}
To continuously improve the quality of the box bank, we implement an iterative refinement mechanism that selectively incorporates newly extracted confident objects. Let $O_{\text{new}}$ denote the set of newly extracted objects and $O_{\text{bank}}$ the set of objects already stored in the box bank. For each new object $o_{\text{new}} \in O_{\text{new}}$, if it overlaps with an existing object $o_{\text{bank}} \in O_{\text{bank}}$, we compare their uncertainty scores, $U(o_{\text{new}})$ and $U(o_{\text{bank}})$, respectively. The object with the lower uncertainty score is retained as follows:

\begin{equation}
o_{\text{retain}} = 
\begin{cases} 
o_{\text{new}}, & \text{if } U(o_{\text{new}}) < U(o_{\text{bank}}) \\
o_{\text{bank}}, & \text{otherwise}
\end{cases}
\end{equation}
This ensures that only higher-confidence objects are kept in the box bank, minimizing the risk of introducing noisy or uncertain objects into the training data.

Despite retaining objects with lower uncertainty scores, errors may still occur in uncertainty estimation, potentially leading to the inclusion of mislabeled or erroneous objects. To mitigate this risk, we introduce a deletion mechanism. After each training iteration, when new confident boxes are extracted, the newly added objects are compared to existing objects in the box bank. The deletion mechanism checks for overlaps between objects, using the Intersection over Union (IoU) metric:

\begin{equation}
\text{IoU}(o_{\text{new}}, o_{\text{bank}}) < \tau_{\text{overlap}}.
\end{equation}
If the overlap is below a specified threshold $\tau_{\text{overlap}}$ for all newly extracted objects, those objects are deleted.

Additionally, mislabeled or erroneous objects are identified and removed by evaluating their performance in subsequent training iterations. This helps ensure that the objects in the box bank remain reliable for training purposes. The iterative process of adding, comparing, and refining the box bank improves its overall quality. This ensures that the uncertainty estimation becomes more reliable over time, formally expressed as:

\begin{equation}
O_{\text{bank}}^{t+1} = \left( O_{\text{bank}}^{t} \setminus O_{\text{remove}} \right) \cup O_{\text{new}},
\end{equation}
where $O_{\text{bank}}^{t}$ is the set of objects in the box bank at iteration $t$, and $O_{\text{remove}}$ is the set of objects identified for deletion during that iteration. Through this iterative refinement, the box bank evolves to contain higher-quality pseudo-labels for training.

\subsubsection{Pseudo Scene Formation}
In typical training scenarios, scenes often contain both confident and unconfident objects. To enhance the model's focus on confident objects while minimizing the impact of unconfident ones, we draw inspiration from the Reliable Background Mining Module in~\citep{ss3d}, which excludes point clouds associated with potential object detections. This method proves highly effective for our task, where we aim to train only on confident objects.
Furthermore, we establish a relatively high threshold to preserve background point clouds that are prone to be falsely identified as positive detections. 
We then construct Pseudo Scenes by merging point clouds from the selected confident boxes in the box bank with these background scenes. These Pseudo Scenes are composed solely of confident objects, excluding any unconfident ones from the training data. This strategy ensures that the pre-training dataset is fine-tuned to help the model make stable and reliable predictions, providing a solid foundation for the active learning process.

\begin{algorithm}[ht]
	\caption{Collaborative Active Learning Algorithm}
	\label{code:cal}
	\textbf{Input}: 
	\begin{itemize}
	    \item Labeled data $D_{l} = \{S_{li}\}_{i=1}^{N_{l}}$
	    \item Unlabeled data $D_{u} = \{S_{ui}\}_{i=1}^{N_{u}}$
	    \item Budget $b$ for selecting new samples
	    \item Class set $C$
	    \item Similarity threshold $T_{sim}$
	    \item Class upper limits $U(c)$ for each class $c \in C$
	    \item Weight adjustment factor $S(c)$ for each class $c \in C$
	\end{itemize}
	\textbf{Output}: Selected data $\Delta D$ for labeling \\
	\textbf{Initialize}: $\Delta D \gets \varnothing$ (selected set of data to be labeled)
	\begin{algorithmic}[1] 
		\STATE Calculate uncertainty of each sample in $D_u$ using $E^2\_Unc$: $\{E_i\}_{i=1}^{N_u}$.
		\STATE Sort $D_u$ in descending order based on uncertainty values $\{E_i\}$.
		\STATE Initialize $idx \gets 0$ and $Box(c) \gets 0$ (the count of boxes for each class $c \in C$).
		\WHILE{$Num_{box}(\Delta D) < b$}
		    \STATE Compute similarity $S_{idx}$ between $D_u(idx)$ and the already selected data $\Delta D$ using $B\_Div$.
		    \IF{$S_{idx} < T_{sim}$}
		        \STATE Add $D_u(idx)$ to $\Delta D$ and remove it from $D_u$.
		        \IF{there exists class $c \in C$ such that $Box(c) < U(c)$ and $Box(c) + Num_{box}(D_u(idx, c)) \geq U(c)$}
		            \STATE Recalculate uncertainties $\{E_i\}_{i=1}^{N_u}$ considering weight adjustment factor $S(c)$.
		            \STATE Resort $D_u$ based on the updated uncertainties.
		            \STATE Set $idx \gets 0$.
		        \ELSE
		            \STATE Set $idx \gets idx+1$.
		        \ENDIF
		        \STATE Update $Box(c)$ for each class based on newly selected data.
		    \ENDIF
		\ENDWHILE
	\end{algorithmic}
\end{algorithm}

\subsection{Collaborative Active Learning}
To efficiently identify the most informative samples and achieve better collaboration with the semi-supervised pre-training stage, we propose a novel Collaborative Active Learning (CAL) approach, which simultaneously incorporates considerations of uncertainty and diversity. For uncertainty, we devise Ensemble-based Entropy Uncertainty ($E^2\_Unc$). In terms of diversity, our approach includes Box-level Diversity ($B\_Div$) and Class Balance Sampling (CBS).   We also present the completed pseudo-code for our active learning process, as shown in Algorithm \ref{code:cal}.

\subsubsection{Ensemble-based Entropy Uncertainty}
We use entropy to measure the uncertainty of each predicted box. 
Considering the collective influence of all the boxes, the overall uncertainty of the entire scene is represented by calculating the average entropy. This approach enables us to capture the overall uncertainty and make informed decisions based on the entropy measure.

Specifically,  the uncertainty for a point cloud scene $S$ is computed as:
\begin{equation}
\begin{aligned}
H(S) = \frac{\sum_{b\in S}\sum_{c\in C}({-p_{bc}\log p_{bc}})}{N_b\times|C|}
\label{al:entropy}
\end{aligned}
\end{equation}
where $b$ represents the predicted boxes, $p_{bc}$ is the predicted class probability of class $c$ for box $b$, $N_b$ is the total number of predicted boxes, and $|C|$ is the number of object classes.

While this entropy-based measure captures uncertainty, it does not fully leverage the potential of the CPSP model. We observe that the CPSP pre-trained model may overlook some confident objects, leading to potential gaps in uncertainty measure. To mitigate this, we propose an ensemble strategy that combines the high-confidence predictions from the normal pre-trained model with all the boxes from the CPSP pre-trained model. Redundant boxes are then removed using the Non-Maximum Suppression (NMS) technique, ensuring more accurate and reliable predictions.

\subsubsection{Box-level Diversity}
Diversity is essential for reducing redundancy in the selected samples. We achieve this by measuring the similarity between boxes using cosine similarity and assigning each box to its closest counterpart. The similarity score for each scene is computed by averaging the cosine similarity between box features from the current scene and features from previously selected scenes.

Formally, let $S_a$ be a scene with box features $F_a = \{f_{a, i} \mid \text{box}_{a_i} \in S_a \}$, and $S = \{S_c\}$ be the set of selected scenes with features $F = \{f_i \mid \text{box}_{i} \in S_c \}$. We calculate the similarity of scene $S_a$ as:

\begin{equation}
Sim_{a} = \frac{1}{|F_a|} \sum_{i=1}^{|F_a|} \max_j \left( \frac{f_{a, i} \cdot f_j}{||f_{a, i}|| \cdot ||f_j||} \right)
\end{equation}
During the sample selection process, if the similarity score between a new sample and previously selected samples exceeds a threshold, the sample is excluded from selection to avoid redundancy. Given the potentially large size of $|F|$, we apply clustering to retain the most representative features.

\subsubsection{Class Balance Sampling}
Outdoor LiDAR scenes pose significant challenges due to the presence of rare classes, which are difficult to sample and annotate. The scarcity of these rare classes makes their annotation disproportionately expensive, especially when they coexist with more frequent classes within the same scene. Additionally, many existing methods~\citep{crb, enms, ppal, kecor} fail to address the inherent difficulty models face in accurately estimating the number of objects in complex outdoor scenes, often resulting in high rates of false positives (FP) and false negatives (FN).

To mitigate these issues, we focus on boxes that are predicted by both the normal pre-trained model and the CPSP pre-trained model. This intersection likely represents true objects, filtering out background noise and reducing false positives, thereby improving the reliability of the sampling process. To further address the class imbalance problem, we propose a class balance algorithm that dynamically adjusts the sampling process by setting an upper limit \( U_c \) for the number of objects selected from each class. If the number of objects from a class exceeds this limit, the weight assigned to that class is reduced from 1 to 0.1, encouraging the model to focus on unconfident objects from underrepresented classes. Specifically, the upper limits for each class \( c \) are determined as:

\begin{equation}
U_c = \frac{N_{\text{total}}/ N_c}{\sum_{i=1}^{C} N_{\text{total}}/ N_i} \cdot B
\end{equation}
where \( N_c \) is the number of labeled samples for class \( c \), \( N_{\text{total}} \) is the total number of labeled samples, \( N_i \) is the number of labeled samples for class \( i \), \( C \) is the total number of classes, and \( B \) is the total number of samples to be selected. To prevent unrealistically small or negative values, a minimum threshold is applied to \( U_c \). This ensures that the model maintains balanced sampling by prioritizing underrepresented classes while avoiding over-sampling of more frequent ones.

This approach ensures rare classes are well-represented and prevents over-sampling of easier classes, achieving a balanced class distribution and improving model robustness.

\section{Experimental Results and Analysis}

\subsection{Datasets and Evaluation Metrics}

\subsubsection{KITTI Dataset}
We conducted evaluations of our methods on the KITTI 3D detection benchmark~\citep{kitti}, using the standard train split comprising 3,712 samples and the validation split containing 3,769 samples~\citep{shi2020pv}. 
In our semi-supervised active learning framework, we initially trained an initial model using randomly selected frames consisting of approximately  200 boxes. Subsequently, we leveraged the remaining unlabeled training data for further model refinement. During active learning, we specifically selected frames that contained around 150 boxes for effective training. The total number of labeled boxes in our approach is approximately 350 boxes, which accounts for less than 2\% of the total boxes present in the KITTI train split. For evaluation, we calculate the mean average precision (mAP) at 40 recall positions for the Car, Pedestrian (Ped), and Cyclist (Cyc), employing 3D IoU thresholds of 0.7, 0.5, and 0.5, respectively, across different difficulty levels: easy, moderate (mod), and hard.  

\begin{figure*}[ht]
\centering
\includegraphics[width=1.0\textwidth]{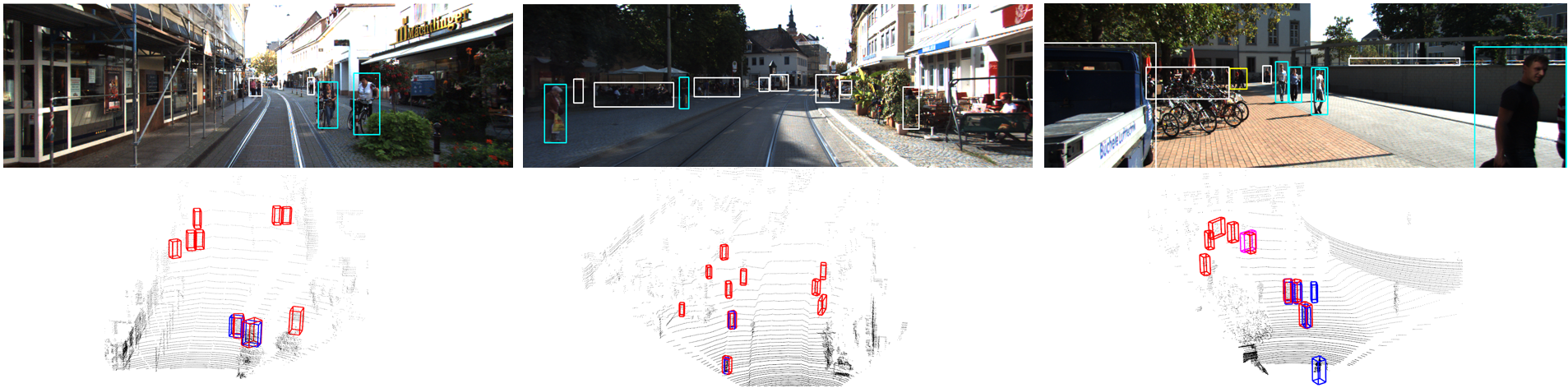}
\caption{
This figure showcases the visualization of selected samples without dropping the ``DontCare" cases. It displays the ground-truth (GT) boxes in green, the predicted boxes in red, and the ``DontCare" areas in white. Each scene is presented through a 2D image and a point cloud representation. In the 2D images, both the GT boxes and the ``DontCare" areas are visualized, while in the point cloud scenes, both the GT and predicted boxes are visualized.
}
\label{fig:dontcare}
\end{figure*}

A significant portion of the dataset contains ``DontCare" regions, as illustrated in Fig.~\ref{fig:dontcare}. These regions often overlap with challenging objects that contribute significantly to uncertainty but are unlabeled, making them unsuitable for active learning. To mitigate this, we project the predicted 3D boxes onto 2D images since ``DontCare" areas only have 2D annotations. If more than two predicted boxes lie within the ``DontCare" regions, we exclude the corresponding frame from the active learning process.

\begin{table*}[ht]
    \centering
    \caption{Comparison of results for various methods under different settings on the KITTI dataset. To ensure a fair comparison, we ensure that all frameworks utilize an identical amount of labeled data. Here, $N_1$ represents the initial box count, while $N_2$ signifies boxes selected through AL.
    }

    \label{tab:kitti_res}
    \resizebox{0.96\textwidth}{!}{
    \begin{tabular}{c|c|ccc|c c c c c c} 
    \toprule
    \multirow{2}{*}{Setting} & \multirow{2}{*}{$ N_1/N_2 $} & \multirow{2}{*}{Pre-train} & \multirow{2}{*}{AL} & \multirow{2}{*}{SSL} & Car\_mod & Ped\_mod & Cyc\_mod & Avg\_easy & Avg\_mod & Avg\_hard \\ 
    & & & & & mAP & mAP & mAP & mAP & mAP & mAP \\ 
    
    \midrule
    \multirow{5}{*}{AL} & \multirow{5}{*}{$ 200/150 $}
    & Normal & Random & $-$ & 74.5 & 37.8  & 44.1 & 67.4 & 52.1 & 47.7  \\ 
    & & Normal & Entropy & $-$ & 73.6 & 48.2  & 51.9 & 71.4 & 57.9 & 53.3  \\ 
    & & Normal & PPAL & $-$ & 74.2 & 41.6  & 46.9 & 66.7 & 54.2 & 49.3  \\ 
    & & Normal & CRB & $-$ & 73.3 & 45.3  & 47.4 & 68.8 & 55.3 & 50.8  \\
    & & Normal & KECOR & $-$ & 73.2 & 46.7  & 48.2 & 69.7 & 56.0 & 51.3  \\

    \midrule
    \multirow{6}{*}{AL+SSL} & \multirow{6}{*}{$200/150$}
    & Normal & Random & HSSDA & 78.8 & 54.1  & 59.9 & 77.1 & 64.3 & 59.7  \\
    & & Normal & Entropy & HSSDA & 79.3 & 59.1  & 64.6 & 79.1 & 67.7 & 62.2  \\
    & & Normal  & PPAL & HSSDA & 80.0 & 56.1  & 66.2 & 79.7 & 67.4 & 61.8  \\
    & & Normal & CRB & HSSDA & 79.0 & 58.7  & 63.9 & 78.7 & 67.2 & 62.7  \\
    & & Normal & KECOR & HSSDA & 79.2 & 59.5  & 64.9 & 80.3 & 67.9 & 63.1  \\ 
    & & Normal & CAL & HSSDA & 80.6 & 60.2  & 67.7 & 81.5 & 69.5 & 64.5  \\ 
    
    \midrule
    \multirow{6}{*}{$SSL_p+AL+SSL$} & \multirow{6}{*}{$200/150$}
    & 3DIoUMatch & Entropy & HSSDA & 78.1 & 57.3  & 64.4 & 80.1 & 66.6 & 61.5  \\
    & & 3DIoUMatch & CAL & HSSDA & \textbf{80.8} & 57.1  & 65.9 & 79.9 & 67.9 & 63.1  \\

    & & Joint3D & CAL & HSSDA & 78.5  & 58.9   & 70.1  & 80.7  & 69.1  & 64.1   \\
    & & NAL & CAL & HSSDA & 79.8  & 59.3  & 69.9  & 81.3 & 69.6  & 64.6  \\
    
    & & HSSDA & Entropy & HSSDA & 78.8 & 52.3  & 68.2 & 79.9 & 66.4 & 62.0  \\
    & & HSSDA & CAL & HSSDA & 79.8 & 59.6  & 66.2 & 80.8 & 68.5 & 63.9  \\

    \midrule
    \multirow{5}{*}{$S-SSAL$} & \multirow{5}{*}{$200/150$}
    & CPSP & Entropy & HSSDA & 79.5 & 57.5  & 68.0 & 80.1 & 68.3 & 62.9  \\
    & & CPSP & PPAL & HSSDA & 79.9 & 55.8  & 68.1 & 80.9 & 67.9 & 62.6  \\
    & & CPSP & CRB & HSSDA & 79.1 & 56.9  & 65.4 & 78.7 & 67.2 & 62.8 \\
    & & CPSP & KECOR & HSSDA &  79.0 & 60.8  & 64.5 & 80.6 & 68.1 & 63.3 \\ 
    & & CPSP & CAL & HSSDA & 79.5 & \textbf{61.2}  & \textbf{70.7} & \textbf{81.8} & \textbf{70.5} & \textbf{65.1}  \\ 
  
    \midrule
    \multirow{1}{*}{$Full$} & \multirow{1}{*}{$-/-$}
    & $-$ & $-$  & $-$ & 84.6 & 59.6  & 72.2 & 82.7 & 72.2 & 68.5  \\
    
    \bottomrule 
    \end{tabular}
}
\end{table*}

\subsubsection{Waymo Dataset}
We conducted evaluations of our methods on the Waymo dataset~\citep{waymo}, a widely used benchmark in autonomous driving. It offers diverse real-world driving scenarios with high-resolution sensor data. The dataset comprises 798 training sequences and 202 validation sequences. Notably, the annotations provide a full 360° field of view. Additionally, the prediction results are categorized into LEVEL 1 and LEVEL 2 for 3D objects based on the presence of more than five LiDAR points and one LiDAR point, respectively. 
To optimize efficiency, we adopted a time-saving approach by setting a sample interval of 20 from the training set to generate a pool of frames. From this pool, we selected frames for our divided datasets.
Similar to our approach in the KITTI dataset, we employed a similar strategy for the Waymo dataset. In the initial stage, we randomly sampled frames with approximately 5000 boxes, and in the active learning stage, we again selected frames with around 5000 boxes. The total number of boxes, which amounts to 10,000, is less than 1\% of the total boxes present in the Waymo train set. For evaluation, we use mean average precision (mAP) for Vehicle (Veh), Pedestrian (Ped), and Cyclist (Cyc) in LEVEL\_1 (L1) and LEVEL\_2 (L2), along with average mAP and heading accuracy weighted AP (mAPH).

\subsection{Implementation Details}
As noted in \citep{boxal}, object detection performance is closely tied to the number of boxes, so for a fair comparison with other methods, we maintain a fixed number of boxes rather than frames. We use PV-RCNN~\citep{shi2020pv}, a widely recognized model in active and semi-supervised learning, as the detector within our semi-supervised active learning framework.

During the temporary model updating phase, we use the same initial labeled and unlabeled data across all methods to ensure fair comparison when leveraging pre-training. In the final model evaluation, we randomly initialize the model to assess the performance gains from selecting better data during active learning. In the Confident Object Extraction module, K-Means clustering is applied, with 20 centers for the KITTI dataset and 50 centers for the Waymo dataset. The similarity threshold ($T_{sim}$) in the $B\_Div$ module is set to 0.9. Due to the limited data, model convergence can be challenging. To address this, we extend training iterations by repeating the data 5 times for KITTI and 15 times for Waymo to ensure sufficient training duration. Other settings, such as learning rate, optimizer, and scheduler, follow those outlined in \citep{shi2020pv} and \citep{openpcdet2020}.

\subsection{Results on KITTI}

\begin{table*}[ht]
    \centering
    \caption{Comparing results across different settings on the Waymo dataset. $N_1$ represents the initial box count, while $N_2$ signifies boxes selected through AL. }

    \label{tab:waymo_res}
    \resizebox{0.96\textwidth}{!}{
    \begin{tabular}{c|c|ccc|c c c | c c} 
    \toprule
    \multirow{2}{*}{Setting} & \multirow{2}{*}{$ N_1/N_2 $} & \multirow{2}{*}{Pre-train} & \multirow{2}{*}{AL } & \multirow{2}{*}{SSL} & Veh(L1/L2) & Ped(L1/L2) & Cyc(L1/L2) & \multicolumn{2}{c}{Avg(L1/L2)}  \\ 
    & & & & & mAP & mAP & mAP & mAP & mAPH \\ 
    
    \midrule
    \multirow{4}{*}{AL} & \multirow{4}{*}{$ 5000/5000 $}
    & Normal & Random & $-$ & 62.9/54.8 & 59.6/51.0  & 41.3/39.8 & 54.6/48.6 & 37.9/33.6  \\ 
    & & Normal & Entropy & $-$ & 61.1/53.2 & 60.6/51.9  & 50.1/48.5 & 57.3/51.2 & 40.4/35.9  \\ 
    & & Normal & CRB & $-$ & 62.7/54.4 & 56.6/48.4  & 54.6/52.7 & 57.9/51.8 & 38.8/34.4  \\ 
    & & Normal & KECOR & $-$ & 61.8/53.5 & 57.1/49.0 & 52.1/50.2  & 57.0/50.8 & 39.1/34.5   \\

    \midrule
    \multirow{5}{*}{AL+SSL} & \multirow{5}{*}{$5000/5000$}
    & Normal & Random & CPSP & 63.1/54.8 & 59.4/50.0  & 46.1/45.3 & 56.2/50.0 & 40.3/36.8  \\
    & & Normal & Entropy & CPSP & 61.5/53.5 & 60.0/51.5  & 54.9/53.0 & 58.8/52.7 & 43.4/38.9  \\
    & & Normal & CRB & CPSP & 63.2/54.6 & 57.2/48.9  & 56.9/54.2 & 59.1/52.6  & 42.1/37.7 \\ 
    & & Normal & KECOR & CPSP & 62.5/53.8 & 57.5/49.7 & 56.1/53.6  & 58.7/52.4 & 42.3/38.1  \\ 
    & & Normal & CAL & CPSP & 62.2/54.3 & 61.7/53.0 & 55.4/53.5 & 59.8/53.6 &  45.1/40.6  \\ 
     
    \midrule
    \multirow{3}{*}{$SSL_p+AL+SSL$} & \multirow{3}{*}{$5000/5000$}
    & 3DIoUMatch & CAL & CPSP & 63.1/54.7 & 60.9/52.7  & 53.3/51.6 & 59.1/53.0  & 43.9/39.2 \\
    & & Joint3D & CAL & CPSP & 61.5/53.9  & 59.7/51.3   & 53.2/51.8  & 58.1/52.3  & 41.8/37.2    \\
    & & NAL & CAL & CPSP & 62.6/54.3  & 60.1/51.7  & 55.1/53.1  & 59.3/53.1 & 42.0/38.1  \\

    \midrule
    \multirow{4}{*}{$S-SSAL$} & \multirow{4}{*}{$5000/5000$}
    & CPSP & Entropy & CPSP & \textbf{64.2}/\textbf{56.2} & 60.3/50.8  & 53.4/51.5 & 59.3/52.8  & 44.1/40.7 \\
    & & CPSP & CRB & CPSP & 63.9/55.0 & 59.1/49.1  & 53.8/51.8 & 58.9/52.0 & 43.0/39.0 \\
    & & CPSP & KECOR & CPSP & 63.0/54.5 & 60.2/50.6 & 54.3/52.4  & 59.2/52.5 & 44.6/41.1  \\ 
    & & CPSP & CAL & CPSP & 62.8/54.8 & \textbf{62.8}/\textbf{54.1}  & \textbf{57.3}/\textbf{55.3} & \textbf{61.0}/\textbf{54.7} & \textbf{46.5}/\textbf{42.4}  \\ 
    
    \midrule
    \multirow{1}{*}{$Full$} & \multirow{1}{*}{$-/-$}
    & $-$ & $-$  & $-$ & 75.4/67.4 & 72.0/63.7  & 65.9/63.4 & 71.1/64.8 & 66.7/60.9  \\
    
    \bottomrule 
    \end{tabular}
}
\end{table*}

We conduct experimental evaluation on different settings: the Active Learning (AL) framework, Semi-Supervised Active Learning (AL+SSL) framework, Pretrain-based Semi-Supervised Active Learning ($SSL_P+AL+SSL$) framework, our Synergistic Semi-Supervised Active Learning framework (S-SSAL), and full-labeled (Full) results. Among all these frameworks, they share a similar pattern. In the stage of temporary model updating(TMU), different pre-train methods are adopted like normal pre-train(training on labeled data only), 3DIoUMatch pre-train~\citep{3dioumatch}, Joint3D pre-train~\citep{joint3dalssl}, NAL pre-train~\citep{Notalllabels} and our CPSP pre-train. 
In the stage of unlabeled sample selection(USS), different active learning methods are used, like Entropy, CRB~\citep{crb}, KECOR~\citep{kecor}, PPAL~\citep{ppal}, and our CAL.
For the final model delivering stage(FMD), we leverage HSSDA~\citep{hssda} due to its demonstrated good performance.
To improve result reliability in the limited KITTI dataset, we ran three times with different seeds and averaged the performance across them.

As shown in Table~\ref{tab:kitti_res}, our S-SSAL framework outperforms all other approaches in average mAP, with notable improvements in challenging classes such as Pedestrian and Cyclist. When comparing pre-training methods while keeping the AL methods fixed, our CPSP pre-training consistently delivers superior performance across all classes. In particular, traditional SSL approaches, such as 3DIoUMatch and HSSDA, tend to negatively impact AL performance, as shown by the significant decline in mAP across different difficulty levels. This highlights the inherent conflict between SSL and AL when traditional SSL methods are used, as they fail to improve uncertainty estimation and object detection in a meaningful way. In contrast, CPSP pre-training enhances the performance of nearly all AL methods, resolving this conflict by improving model calibration and uncertainty estimation. 

Furthermore, when pre-training methods are fixed and different AL strategies are employed, our CAL method consistently demonstrates superior performance across various pre-training configurations. This indicates that CAL is highly effective regardless of the pre-training method used, ensuring better uncertainty estimation and addressing class imbalance, especially for rare classes like Cyclist. The results clearly show that S-SSAL, with its CPSP pre-training and CAL, provides a more balanced and robust approach to 3D object detection, outperforming other frameworks and showing the critical importance of our proposed solutions.

Although some work, like 3DIoUMatch,  outperforms our method in the Car category on KITTI, this primarily stems from class imbalance. Specifically, in the setting with 350 labeled objects, 3DIoUMatch samples 264 Cars, 63 Pedestrians, and 21 Cyclists, while our method samples 181 Cars, 128 Pedestrians, and 38 Cyclists. This gives 3DIoUMatch an advantage in the Car category due to over-sampling but results in weaker performance for other classes. We believe that their sampling distribution is not ideal for object detection tasks, where all classes are important. When we adjusted our sampling strategy by assigning significantly higher weights to the Car class (260 Cars), our method achieved an 81.6 mAP in Car, surpassing 3DIoUMatch (80.8). This result demonstrates that our method can achieve competitive performance in individual categories while maintaining balanced performance across all classes.

\subsection{Results on Waymo}

For the Waymo dataset, we utilize CPSP in the FMD stage to accelerate the training process. As shown in Table~\ref{tab:waymo_res}, our S-SSAL framework continues to deliver strong performance, even with a large amount of data, as observed in the Waymo dataset.

When comparing pre-training methods under the same active learning approach, CPSP pre-training outperforms other methods, achieving a 1.2\% improvement in average mAP. This improvement is particularly evident in the Pedestrian category, with a 1.1\% gain in mAP, and the Cyclist category, where we observe a 1.9\% improvement over other pre-training methods using CAL. Specifically, S-SSAL with CPSP pre-training achieves better performance than baseline methods like Normal + Random, Normal + Entropy, Normal + CRB, and Normal + KECOR, as well as traditional SSL approaches like 3DIoUMatch and Joint3D, demonstrating its superior handling of uncertainty and challenging classes.

\begin{table}[t]
\centering
\caption{Ablation study of different components in CPSP. mAP is calculated under the moderate difficulty level.}
\resizebox{0.95\columnwidth}{!}{
    \begin{tabular}{ccc|ccc|c}
        \hline
        \multicolumn{3}{c|}{CPSP}  & \multicolumn{3}{c|}{3D Detection} & \multirow{2}{*}{mAP} \\ 
         $ITER$ & $DEL$ & $FP$  & Car & Ped. & Cyc. &  \\ 
         \hline
        - & -  & - & 78.9 & 56.3  & 66.6 & 67.3  \\ 
        - & -  & \checkmark & 79.0 & 56.9  & 67.1 & 67.7  \\ 
        \checkmark & - & -  & 79.0 & 55.5  & 64.4 & 66.3  \\
        \checkmark & \checkmark & - & \textbf{79.1} & 57.1  & 64.8  & 67.1 \\
        \checkmark & - & \checkmark & 79.0 & 57.0  & 67.8 & 68.0 \\
        \checkmark & \checkmark & \checkmark  & \textbf{79.1} & \textbf{58.7}  & \textbf{67.9} & \textbf{68.6} \\ 
        \hline
    \end{tabular}
}
\label{tab:abl_cpc}
\end{table}

Furthermore, the CAL method we propose shows superior performance across all pre-training methods, consistently outperforming alternatives by at least 1.6\% mAP. This demonstrates CAL’s effectiveness in improving model robustness, particularly for rare classes like Cyclist. This improvement is especially evident in rare categories like Cyclist, which are often underrepresented in other frameworks. By targeting these underrepresented classes, CAL helps mitigate class imbalance, enhancing overall model performance.

Additionally, our methods lead to exceptional improvements in heading estimation accuracy, reflecting the overall effectiveness of S-SSAL in enhancing both object detection and orientation prediction. This highlights the robust capabilities of our approach in handling complex 3D object detection tasks.

\begin{figure}[ht]
\centering
\includegraphics[width=0.9\columnwidth]{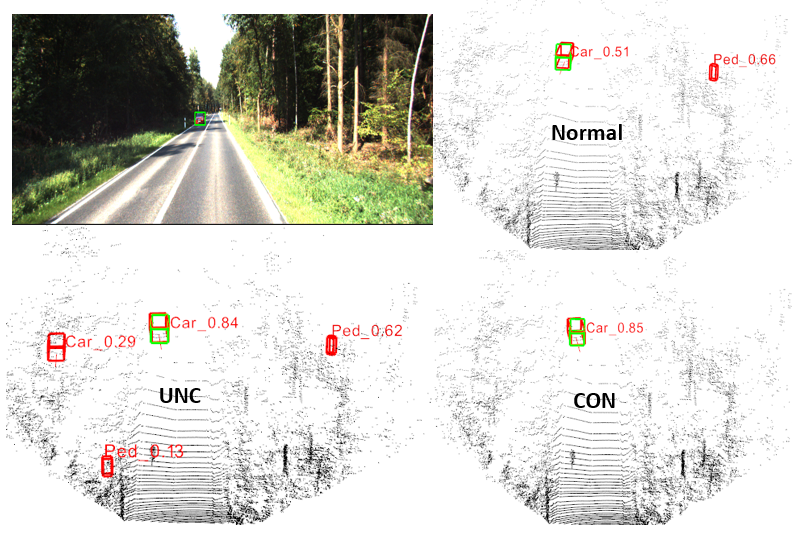} 
\caption{
Visualization of different pre-training methods. Green represents GT boxes, while red indicates predicted boxes, with numbers denoting confidence scores. In Normal, an unconfident car and a false positive pedestrian (tree) are detected. UNC is impacted by noise from unconfident labels, leading to more false positives. In contrast, CON learns the car effectively and eliminates the false positives.
}
\label{fig:unc_cer}
\end{figure}

\subsection{Ablation study}
In this section, we present a series of ablation studies to analyze the effect of our proposed strategies in S-SSAL. 

\subsubsection{Ablation study of Confident Object Extraction (CPSP)}
Table~\ref{tab:abl_cpc} presents the results of the ablation study for CPSP, illustrating the impact of different components—iteration mechanism (ITER), deletion mechanism (DEL), and false-positive (FP)—on model performance. The results indicate that each component contributes positively to the overall performance.

Specifically, adding FP greatly enhances Pedestrian detection, improving mAP from 56.3 to 56.9. This improvement is consistent with the fact that detecting Pedestrian often generates more false positives (e.g., trees or background elements), necessitating the inclusion of additional false positives to help refine the model’s ability to distinguish between true and false detections.
On its own, ITER leads to a decrease in performance, with mAP dropping to 66.3, mainly due to the noise spread during the iterative process. This highlights the importance of the other components in mitigating the negative effects of noisy data. However, when ITER is combined with the DEL mechanism, performance improves, as the DEL mechanism effectively removes unwanted or erroneous objects from the box bank, enhancing the model’s overall performance.
Finally, when all three mechanisms (ITER, DEL, and FP) are used together, the model achieves its best performance, reaching 68.6 mAP, demonstrating improved detection across all classes.

Furthermore, Confident objects provide crucial information with minimal noise, while unconfident ones introduce numerous incorrect pseudo-labels that mislead the model. 
We utilize two types of pre-training in our approach CPSP: UNC (Unconfident) and CON (Confident). The visualization results can be seen in Fig.~\ref{fig:unc_cer}.
In UNC pre-training, objects with a low number of clustering centers (2) are filtered, allowing us to include more unconfident objects in the model training process. In CON pre-training, we train confident objects using the same methods but with a higher number of clustering centers (20). This approach allows us to focus on objects that exhibit a higher level of certainty and reliability during the training process.

Additional results are shown in Table~\ref{tab:abl_cpc_hyperparam}, where we test different numbers of clustering centers (2, 5, 10, 20, and 50). These results highlight the impact of the number of centers on performance, with 20 centers yielding the highest mAP at 68.6. Using fewer centers (2, 5, or 10) introduces noise and reduces the model’s ability to accurately measure uncertainty. Conversely, using more centers (50) fails to effectively leverage the unlabeled data, limiting the model's optimization and performance improvement.

\begin{table}[t]
\centering
\caption{Ablation study of different components in CPSP. mAP is calculated under the moderate difficulty level.}
\resizebox{0.88\columnwidth}{!}{
    \begin{tabular}{c|ccc|c}
        \hline
        \multirow{2}{*}{Numbers}  & \multicolumn{3}{c|}{3D Detection} & \multirow{2}{*}{mAP} \\ 
         & Car & Ped. & Cyc. &  \\ 
         \hline
         2 & 78.7 & 56.1  & 65.3 &  66.7 \\ 
         5  & 79.0 & 56.8  & 65.9 & 67.2  \\
         10  & 78.9 & 57.6  & 66.1 & 67.5  \\
         20 & \textbf{79.1} & \textbf{58.7}  & \textbf{67.9} & \textbf{68.6} \\
         50   & \textbf{79.1} & 58.2  & 67.1 & 68.1 \\ 
        \hline
    \end{tabular}
}
\label{tab:abl_cpc_hyperparam}
\end{table}

\subsubsection{Ablation study of Collaborative Active Learning (CAL)}
The ablation study, as shown in Table~\ref{tab:abl_cal}, highlights the importance of the uncertainty measure, class balance methods, and diversity methods in CAL for improving performance.

The CBS method significantly addresses class imbalances, particularly for harder classes like Pedestrian (improving mAP by 1.7\%) and Cyclist (improving mAP by 10.1\%), which results in better performance on these challenging categories.
The $E^2\_Unc$ component plays a critical role in active learning by selecting informative samples, enabling the model to focus on challenging instances. This leads to an overall 2.4\% mAP improvement.
Additionally, the $B\_Div$ method reduces redundancy in the selected samples, allowing the model to capture a broader range of object variations. This further enhances detection capabilities, contributing to the overall improvement.

By incorporating these CAL components, the overall semi-supervised active learning framework becomes more effective, resulting in better performance in 3D object detection.

\begin{table}[h]
\centering
\caption{Ablation study of different components in CAL. mAP is calculated under the moderate difficulty level.}
\resizebox{0.99\columnwidth}{!}{
    \begin{tabular}{ccc|ccc|c}
        \hline
        \multicolumn{3}{c|}{CAL}  & \multicolumn{3}{c|}{3D Detection} & \multirow{2}{*}{mAP} \\ 
         $CBS$ & $E^2\_Unc$ & $B\_Div$  & Car & Ped. & Cyc. &   \\ 
         \hline
        - & -  & - & 78.0 &  51.8 & 53.3 & 61.3  \\ 
        \checkmark & - & -  & 78.6 & 53.5  & 63.4 & 65.2  \\
        \checkmark & \checkmark & - & 79.2  & 56.9 & 66.7  & 67.6 \\
        \checkmark & - & \checkmark & \textbf{79.3}  & 54.8 & 66.9  & 67.0 \\
        \checkmark & \checkmark & \checkmark  & 79.1 & \textbf{58.7}  & \textbf{67.9} & \textbf{68.6} \\ 
        \hline
    \end{tabular}
}
\label{tab:abl_cal}
\end{table}

\begin{figure}[h]
\centering
\includegraphics[width=0.99\columnwidth]{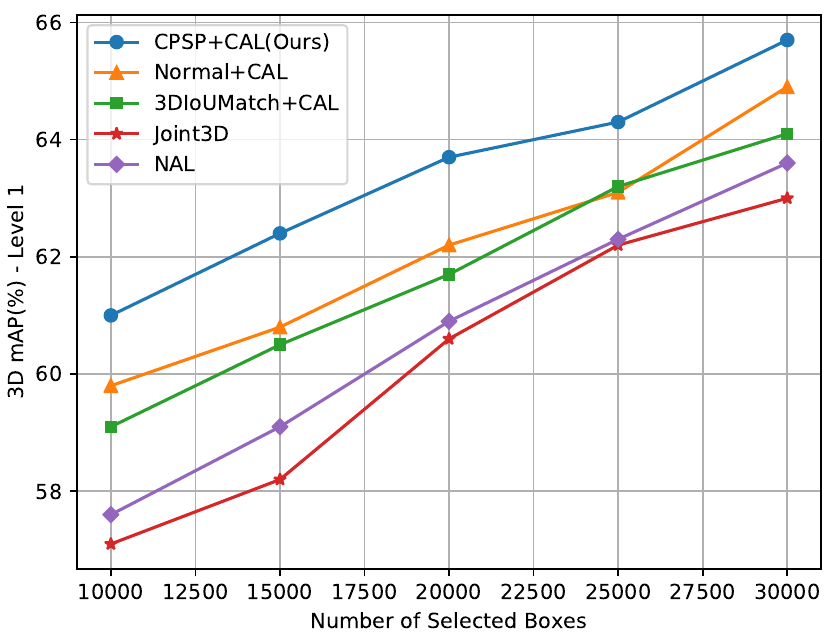} 
\caption{
Performance of different methods in multiple rounds on Waymo Datasets.
}
\label{fig:mrounds}
\end{figure}

\begin{table*}[h]
\centering
\caption{Results of single-round and three-round experiments on Waymo, with a total of 20,000 objects.}

\resizebox{0.95\textwidth}{!}{
\begin{tabular}{c|c|c|ccc|c} 
\toprule

Setting & Rounds & Numbers  & Veh(L1/L2) & Ped(L1/L2) & Cyc(L1/L2) & Avg(L1/L2) \\
\hline
CPSP + CAL & 1 & 5000+15000 & 65.3/57.1 & 65.0/56.2  & 58.1/55.9 &62.8/56.3   \\
\hline
CPSP + CAL & 3 & 5000+3$\times$5000  & \textbf{65.8}/\textbf{57.6} & \textbf{66.0}/\textbf{57.2}  & \textbf{59.2}/\textbf{56.8} & \textbf{63.7}/\textbf{57.3}  \\ 
\bottomrule 
\end{tabular}
}
\label{tab:multi_round}
\end{table*}

\begin{table*}[h]
    \centering
    \caption{Comparison of results for various methods on the KITTI dataset, with all frameworks using the same amount of labeled data. $N_1$ denotes the initial number of boxes, and $N_2$ represents boxes selected during active learning.}
    
    \label{tab:kitti_res_cpsp}
    \resizebox{0.95\textwidth}{!}{
    \begin{tabular}{ccc|c|c c c c c c} 
    \toprule
     \multirow{2}{*}{Pre-train} & \multirow{2}{*}{AL} & \multirow{2}{*}{SSL} & \multirow{2}{*}{$ N_1/N_2 $}  & Car\_mod & Ped\_mod & Cyc\_mod & Avg\_easy & Avg\_mod & Avg\_hard \\ 
    & & & & mAP & mAP & mAP & mAP & mAP & mAP \\ 
    
    \midrule
     Normal & CAL & CPSP & \multirow{1}{*}{$200/150$}  & 78.8 & 55.6  & 65.8 & 79.6 & 66.7 & 61.9  \\

    \midrule
     Joint3D & CAL & CPSP & \multirow{3}{*}{$200/150$} & 77.7 & 55.1  & 63.4 & 77.2 & 65.3 & 60.7  \\
    NAL & CAL & CPSP  & & 77.1 & 57.8  & 63.1 & 78.8 & 66.0 & 61.0  \\
    HSSDA & CAL & CPSP  & & 77.2 & 56.7  & 65.3 & 79.1 & 66.4 & 61.5  \\

    \midrule
    CPSP & CAL & CPSP & \multirow{1}{*}{$200/150$} & \textbf{79.1} & \textbf{58.7}  & \textbf{67.9} & \textbf{80.2} & \textbf{68.6} & \textbf{62.8}  \\

    \bottomrule 
    \end{tabular}
}
\end{table*}

\begin{figure*}[!ht]
\centering
\includegraphics[width=1.0\textwidth]{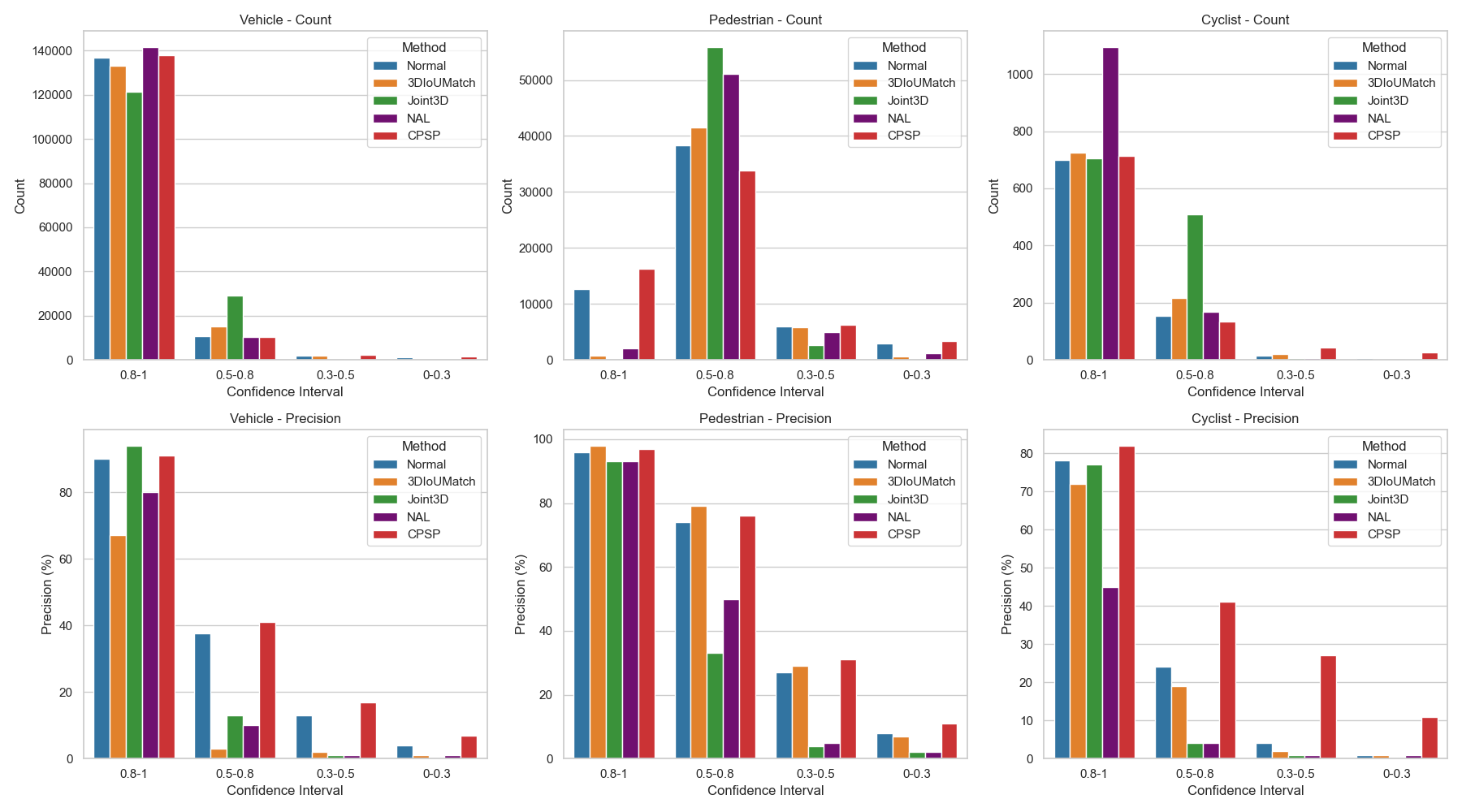} 
\caption{Histogram illustrating the distribution of Precision and Count across different pre-training methods for three object classes. The figure highlights the comparative performance of each method in terms of prediction accuracy and the number of predictions within specific confidence intervals.}

\label{fig:calib_analysis}
\end{figure*}

\begin{figure*}[h]
\centering
\includegraphics[width=0.9\textwidth]{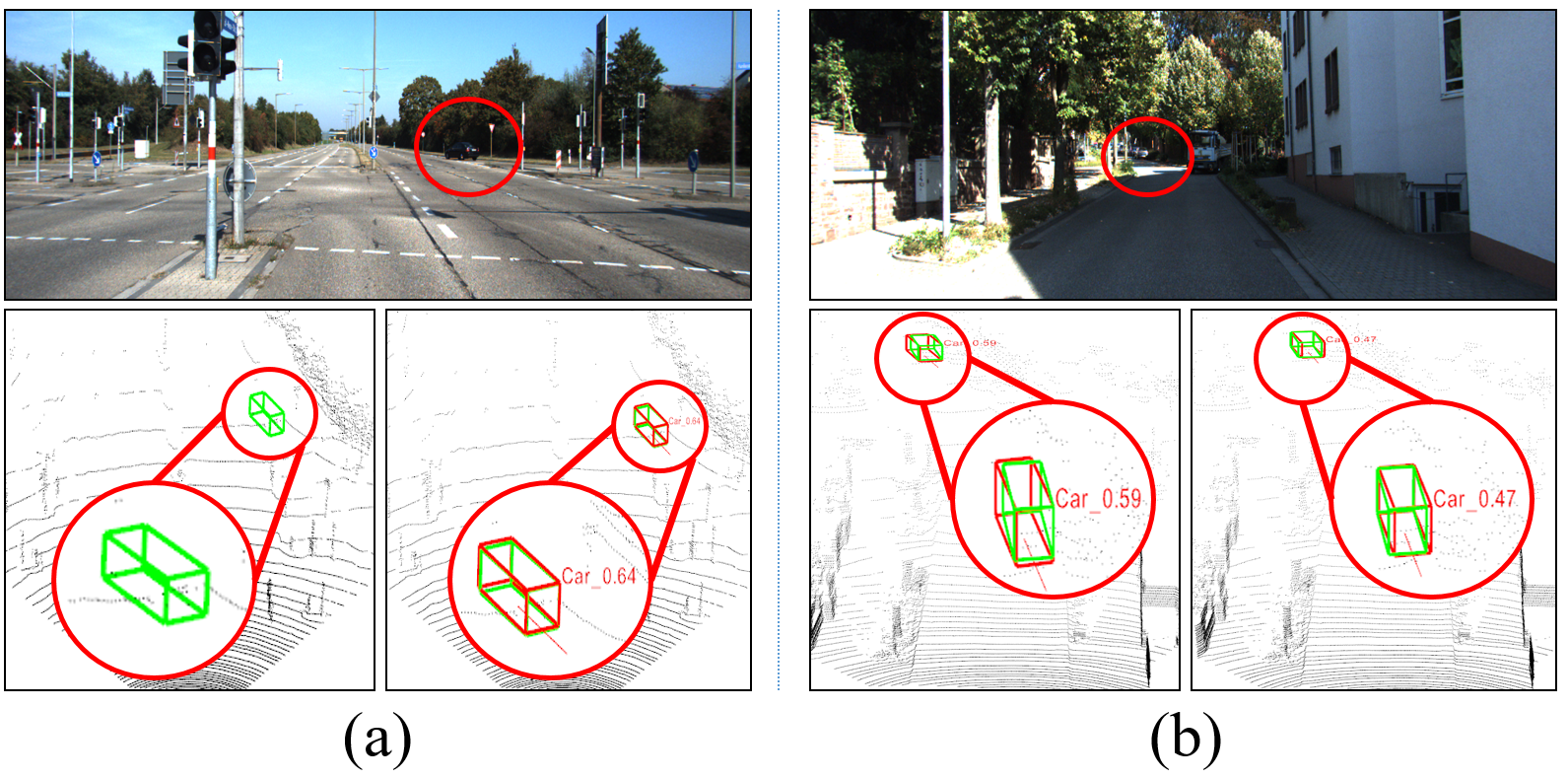} 
\caption{Qualitative results of selected samples. Green boxes represent GT boxes, while the red boxes denote the predicted boxes. We visualize two scenes, one located on the left(a) and the other on the right(b). Each scene is presented with three images: the top image shows the corresponding 2D image, the bottom-left image displays the predicted results from the normal pre-trained model, and the bottom-right image shows the predicted results from the CPSP pre-trained model.}
\label{fig:res_vis}
\end{figure*}

\begin{figure*}[h]
\centering
\includegraphics[width=0.95\textwidth]{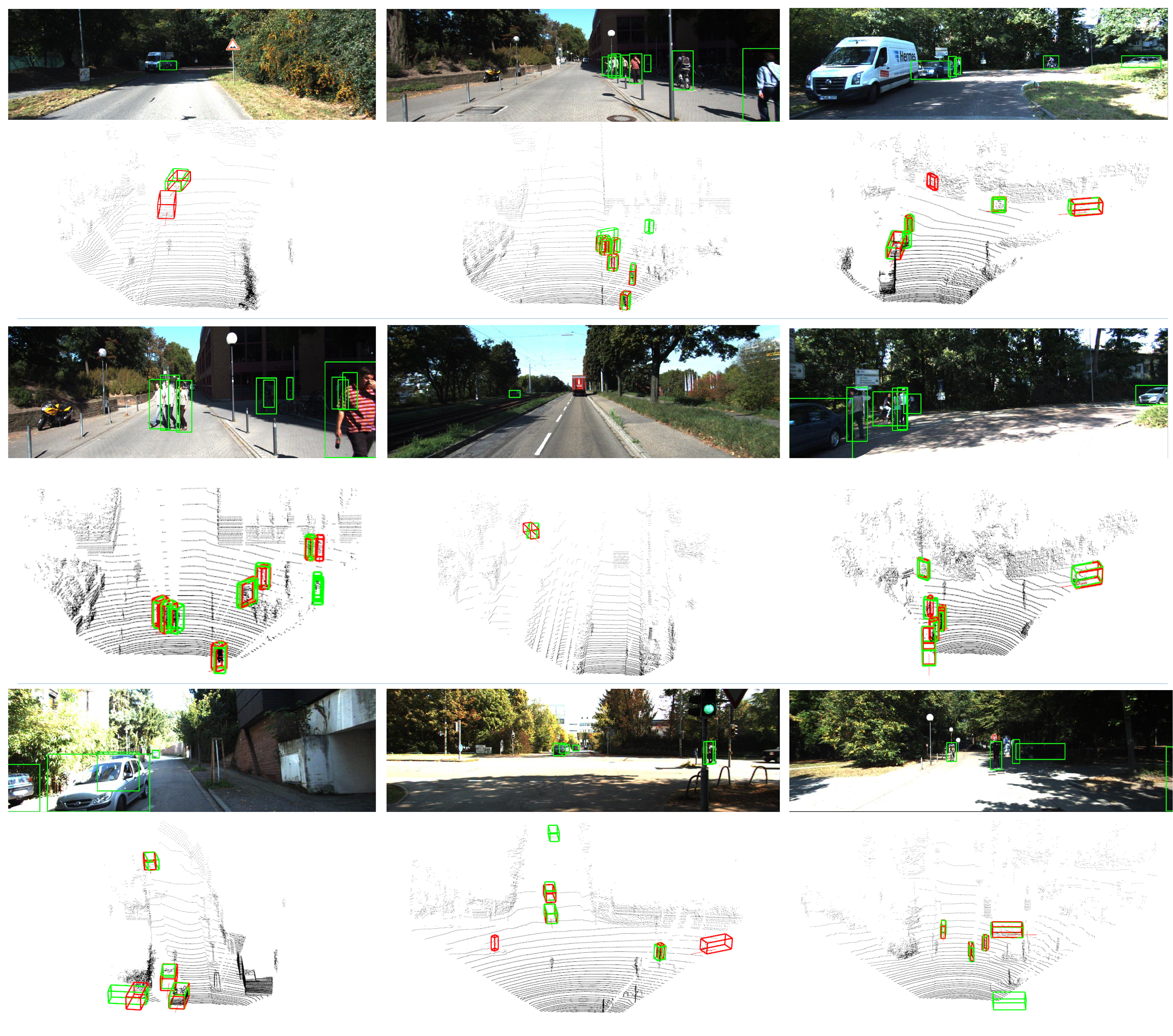} 
\caption{
This figure displays the visualization of selected samples, showcasing the ground-truth (GT) boxes in green and the predicted boxes in red. Each scene is represented by both a 2D image and a point cloud. In the 2D images, only the GT boxes are visualized, while in the point clouds, both the GT and predicted boxes are visualized. 
}
\label{fig:allsamples}
\end{figure*}

\subsubsection{Multiple Rounds of Experiments}
\label{multi_round_exp}
We conducted experiments using both single-round and multi-round approaches to assess their potential for improving performance. Specifically, we performed experiments on the Waymo dataset, utilizing a total of 20,000 annotated boxes to compare the outcomes of single-round versus three-round approaches. 
As shown in Table~\ref{tab:multi_round}, increasing the number of rounds while maintaining the same annotation budget results in a 0.9\% improvement in mAP. We attribute this enhancement to the model's improved ability to select better data over multiple rounds, underscoring the positive impact of utilizing multiple rounds in our approach.

In addition, to facilitate a comprehensive comparison of our methods with existing approaches, we employ active learning over multiple rounds. In each round, we annotate 5000 boxes, conducting a total of 5 rounds, which culminate in 30,000 annotated boxes. We compare the following methods in our evaluation: Joint3D~\citep{joint3dalssl}, NAL~\citep{Notalllabels}, 3DIoUMatch~\citep{3dioumatch} combined with our CAL, Normal pre-training combined with CAL, and our CPSP pre-training combined with CAL. For a fair comparison, all methods utilize CPSP in the Final Model Delivering Stage.
As depicted in Fig.~\ref{fig:mrounds}, our method (CPSP + CAL) consistently outperforms all other approaches across every round of active learning, demonstrating substantial performance gains even in the final evaluation round. The sustained advantage of our method highlights its effectiveness in selecting and leveraging superior data throughout the active learning process.

\subsubsection{Different Semi-supervised Schemes}
We replace the SSL methods in Table~\ref{tab:kitti_res} with our Collaborative PseudoScene Pre-training (CPSP) approach in the Final Model Delivering Stage to examine how well our method performs when compared with different pre-train methods. As presented in Table~\ref{tab:kitti_res_cpsp}, the results demonstrate that our CPSP method not only achieves the best performance but also generates a more significant performance gap in comparison to other pre-train methods. The improvement is particularly noticeable across challenging classes such as pedestrians and cyclists. This highlights the robustness of CPSP in handling diverse object detection tasks.

\subsection{Analysis about different pre-training methods.}
To evaluate how well our CPSP pre-trained model aligns with uncertainty-based active learning methods for object detection, we focus on the key aspect: Calibration.

Calibration~\citep{calibration} refers to how accurately the model’s confidence scores reflect the correctness of its predictions. A well-calibrated model is crucial for active learning, as it helps in selecting the most informative samples. We use D-ECE~\citep{caliod} to measure the calibration quality. 
As noted before, the KITTI dataset includes many ``DontCare'' labels, making it challenging to accurately calculate D-ECE scores. Therefore, we conduct our analysis using the Waymo training set. As shown in Table~\ref{tab:calib_recall_1}, our CPSP model achieves strong performance in both D-ECE, better supporting the active learning process. In contrast, other methods perform poorly in D-ECE, making them less effective for active learning. 

\begin{table}[t]
\centering
\caption{D-ECE scores for different pre-train methods on Waymo training set.}
\resizebox{0.95\columnwidth}{!}{
    \begin{tabular}{c|ccc}
        \hline
        \multirow{2}{*}{Pre-train}  & \multicolumn{3}{c}{D-ECE $\downarrow$}   \\
          & Veh & Ped. & Cyc.    \\
        \hline
        Normal   & 0.11 & 0.10 & 0.25   \\ 
        \hline
        3DIoUMatch~\citep{3dioumatch}   & 0.50 & 0.13 & 0.29   \\ 
        \hline
        Joint3D~\citep{joint3dalssl}   & 0.30 & 0.36 & 0.48  \\ 
        \hline
        NAL ~\citep{Notalllabels}  & 0.28 & 0.26 & 0.42   \\ 
        \hline      
        CPSP   & \textbf{0.09} & \textbf{0.08} & \textbf{0.15}   \\ 
        \hline

    \end{tabular}
}

\label{tab:calib_recall_1}
\end{table}

To gain a deeper understanding of why our CPSP pre-training method outperforms other pre-training methods, we further analyze the model's calibration beyond the D-ECE score. Specifically, we divide the confidence scores into four ranges: 0-0.3, 0.3-0.5, 0.5-0.8, and 0.8-1. The scores in the range of 0.3-0.8 represent more uncertain predictions, which are more likely to be selected by active learning, while the range of 0.8-1 corresponds to predictions that are more likely to represent real objects.

As shown in Fig.~\ref{fig:calib_analysis}, our CPSP method demonstrates superior calibration, particularly in the range of 0.3-0.8, which is critical for active learning tasks. The accuracy of predictions within this range is significantly higher compared to other pre-training methods, indicating that CPSP better handles uncertain predictions and reduces overconfidence. Moreover, CPSP consistently outperforms Normal across all three object classes, maintaining both higher precision in high-confidence predictions (in the range of 0.8-1) and better accuracy for uncertain predictions (in the range of 0.3-0.8). This balanced calibration makes CPSP especially effective for active learning, where selecting informative samples from uncertain predictions is crucial for optimizing model performance with limited data.

\subsection{Qualitative Results}
We present visualizations of selected samples in Fig.~\ref{fig:res_vis}. In Fig.~\ref{fig:res_vis}(a), we observe that our CPSP pre-trained model is capable of detecting hard objects that are missed by a model trained with normal pre-training. This highlights the effectiveness of our CPSP approach in discovering challenging objects.
In Fig.~\ref{fig:res_vis}(b), we showcase how our CPSP pre-trained model retains uncertainty for real unconfident boxes. This ability to maintain uncertainty is crucial for effective active learning, enabling the model to focus on challenging examples and improve its performance.

Additional selected samples are shown in Fig.~\ref{fig:allsamples}, which provides a visual representation of challenging instances across various object classes. These samples cover a wide range of scenarios, emphasizing the model's ability to focus on hard examples that require precise detection and localization. The visualization of these challenging samples demonstrates the effectiveness of our active learning strategy in prioritizing difficult-to-detect objects.

\section{Conclusion}
In this paper, we propose a \textbf{S}ynergistic \textbf{S}emi-\textbf{S}upervised \textbf{A}ctive \textbf{L}earning framework, dubbed as S-SSAL, which consists of Collaborative PseudoScene Pre-training (CPSP) and Collaborative Active Learning (CAL), effectively addressing the conflicts between semi-supervised learning and active learning. CPSP utilizes pseudo scenes with confident boxes for model pre-training, while CAL maximizes the benefits of the CPSP pre-trained model to select superior samples. Experimental results on KITTI and Waymo datasets demonstrate that our approach achieves state-of-the-art performance, offering a promising solution for improving 3D object detection through effective integration of semi-supervised and active learning.

\section*{Data Availability Statements}
The KITTI~\citep{kitti}, Waymo~\citep{waymo}databases used in this manuscript are deposited in publicly available repositories respectively: \url{https://www.cvlibs.net/datasets/kitti} and \url{https://waymo.com/open}.

\bibliographystyle{spbasic}      
\bibliography{egbib}   

\end{document}